\documentclass[runningheads]{llncs}
\usepackage{graphicx}
\usepackage[T1,T5]{fontenc}
\usepackage{amsmath,amssymb,amsfonts}
\usepackage{algorithmic}
\usepackage{graphicx}
\usepackage{textcomp}
\usepackage{comment}
\usepackage {bbding}
\begin{document}
\title{Using Artificial Intelligence and IoT for Constructing a Smart Trash Bin} 
%
%
\author{Khang Nhut Lam\Envelope \inst{1} \and
Nguyen Hoang Huynh \and
Nguyen Bao Ngoc \and
To Thi Huynh Nhu \and
Nguyen Thanh Thao \and
Pham Hoang Hao \and
Vo Van Kiet \and
Bui Xuan Huynh \and
Jugal Kalita\inst{2}}
\authorrunning{Lam et al.}
%
\institute{Can Tho University, Can Tho, Vietnam \\  
\email{lnkhang@ctu.edu.vn,\{n.h.huynh2019, nbngoc.grace, sameto98.huynhnhu, nguyenthanhthao9809, haopham6798, vovankiet1108, buixuanhuynha1\}@gmail.com } \and
University of Colorado, Colorado Springs, USA\\
\email{jkalita@uccs.edu}}
\maketitle              
\vspace{-0.4cm}
\begin{abstract}
The research reported in this paper transforms a normal trash bin into a smarter one by applying computer vision technology. With the support of sensor and actuator devices, the trash bin can automatically classify garbage. In particular, a camera on the trash bin takes pictures of trash, then the central processing unit analyzes and makes decisions regarding which bin to drop trash into. The accuracy of our trash bin system achieves 90\%. Besides, our model is connected to the Internet to update the bin status for further management. A mobile application is developed for managing the bin.

\end{abstract}
\vspace{-0.8cm}
\section{Introduction}\vspace{-0.2cm}
Waste management has become an issue of great concern. People are aware of the impact of improper waste disposal practices, but the inconvenience of implementation often leads to unexpected results. Existing studies propose several approaches to help manage trash, protect the environment, and support communities. GirdharSarda et al. \cite{ref_GirdharSarda19} contribute to solving the problem of garbage emission in India by building a smart trash can using IoT. The trash bin has an Arduino connected with an ultrasonic sensor for measuring distance and a moisture sensor for detecting whether the waste is wet or dry. 
Another similar project on automatic trash classification \cite{ref_Chandramohan14} uses electronic and sensors's knowledge. This system classifies metal, moisture, and dry waste. In particular, it uses a parallel resonance impedance sensing mechanism to detect metals and capacitive sensors to distinguish between wet and dry waste. Similarly, Sharanya et al. \cite{ref_Sharanya17} use an Arduino UNO to construct a waste segregator which can classify waste into metallic, wet, and dry. Lopes and Machado \cite{ref_Lopes19} construct an automatic household waste segregator which can detect and classify dry, wet, and metallic waste, and monitor the waste level in the bin. The system consists of a variety of sensors: an ultrasonic sensor is used to detect an object, a metallic sensor is used to check for any metal content in the waste, and a capacitive detector is used to distinguish between dry and wet waste. In addition, the GSM and Arduino are used to send a message to clean the bin if it is full. Mahajan et al. \cite{ref_Mahajan17} build a smart waste management system using different sensors and Raspberry Pi. The ultrasonic  sensor is used to detect the level of garbage in the trash cans, the humidity sensor is used to distinguish between dry and wet waste, the load sensor is used to improve data related to waste level.

Artificial intelligence has been used in garbage classification to improve efficiency. Computer vision technology is used to collect garbage images to classify them into six categories. Thung and Yang \cite{ref_Yang16} manually collect data consisting of 400-500 images for each class. The Support Vector Machines (SVM) models and Convolutional Neural Networks (CNN) are used to classify garbage. Their experiments show that the SVM outperformed the CNN model. Costa et al. \cite{ref_Costa18} experiment with Thung and Yang’s dataset using different methods to find the best performance among pre-trained models, VGG-16, AlexNet\cite{ref_Alex12}, SVM, K-Nearest Neighbor, and Random Forest. Their system achieves up to 93\% accuracy with the VGG-16 model. Sudha et al. \cite{ref_Sudha16} use deep learning to classify biodegradable and non-biodegradable objects in real-time. Tran et al. \cite{ref_Tran20} design a waste management system using IoT, graph theory, and machine learning to predict the probability of waste levels in trash bins. 

Besides the studies about constructing smart trash cans and classifying garbage, several products have been introduced to the market. The new BIN-E\footnote{https://bine.world/} is a smart waste bin that sorts and compresses the recyclables automatically. The Xiaomi smart trash can\footnote{https://xiaomi-mi.com/news-and-actions/review-townew-smart-trash-can/} automatically opens the lid up by using a smart sensor to detect objects, keep the odors sealed in, and put a new bag in place automatically. In Vietnam, the Bridgestone Tire Sales Vietnam Limited Liability Company produces Bridgestone smart garbage bins\footnote{https://www.youtube.com/watch?v=QDwyw9oLC1Q} having the function of classifying and deodorizing garbage. In addition, the water extracted and filtered from organic garbage is used to water two pots of plants on either side of the bin to increase the aesthetics and create a green space around the bin.

The goal of our project is to construct a smart domestic waste classification system. The system can automatically classify garbage and is connected with a mobile application to allow users to perform management operations. Garbage is divided into 2 main types: recyclable and non-recyclable trash. Recyclable garbage includes plastic bottles, cans, paper, and pens; while non-recyclable garbage consists of plastic bags, styrofoam containers, food packets, and plastic glass. This classification method can be changed to serve different agencies or countries depending on their classification standards. We develop a garbage classifier running on a Raspberry Pi. The Raspberry Pi  is responsible for connecting with sensors and actuators, which allow the trash bin to operate automatically. Besides, a Raspberry Pi device with an appropriate configuration can send data to the mobile application. Another important part of our trash system is a mobile application to manage information and track the trash bin status. This application is developed with React-Native and Firebase platforms for back-end development. The rest of this paper describes the trash bin system architecture and the results of the system developed.
\vspace{-0.4cm}
\section{System Components} \vspace{-0.3cm}
The trash bin system comprises the following main modules: Raspberry Pi 3 B+, Raspberry Pi camera V1 5MP module, servo motor MG90S, PIR sensor SR501 module, and the IR Infrared Obstacle Avoidance.

\begin{itemize}
	\item Raspberry Pi\footnote{https://www.raspberrypi.org/products/raspberry-pi-3-model-b-plus/} is a small-sized computer with powerful built-in hardware. It is capable of running the operating system and many applications on it. Raspberry Pi 3 Model B+ is a processor and network connector. 
	The Raspberry Pi plays an important role in our trash bin system.
 	 
	\item The Raspberry Pi camera module V1 5 Megapixel\footnote{https://raspberrypi.vn/san-pham/raspberry-pi-camera-module} acts as the eyes of the entire system by transmitting images taken from daily waste to the Raspberry Pi to analyze. Raspberry Pi camera has high light sensitivity and works well in many different lighting conditions (both indoor and outdoor conditions). 
	
	\item PIR (Passive and InfraRed sensor) HC-SR501 Sensor\footnote{http://www.generationplus.biz/index.php/electronic-parts/pir-hc-sr501} is used to detect the motion of surrounding people and activate the camera and the processor if it senses a person moving.  
	
	\item RC Servo MG90S motor\footnote{https://components101.com/motors/mg90s-metal-gear-servo-motor} is a micro servo motor with metal gear. This small and lightweight servo comes with high output power, thus ideal for strong traction and high durability. The motor is used for moving waste to the right place through the steps of programming control.  
	
	\item IR Infrared Obstacle Avoidance\footnote{https://hshop.vn/products/cam-bien-vat-can-hong-ngoai-v1-2/} is used to detect if the trash bin is full.  
\end{itemize}
 \vspace{-0.8cm}
\section{Data Collection} \vspace{-0.3cm}
While training an image classification model, the size and quality of training data may affect the results. The number of images and their diversity are the two important factors that help the model work well. We collect images from a variety of sources and manually eliminate error images by removing blank, error, or poor quality images.
\begin{itemize}
	\item {Dataset 1}: This dataset is a combination of the images provided by Thung and Yang \cite{ref_Yang16} and images extracted from the Internet. Thung and Yang provide a dataset containing 2,527 images of a variety of types of garbage in the condition of white background and normal light (sunlight or room light). To suit our purpose, we use 24 images of cover cartons and 56 images of metal. In addition, we collect a large amount of data using Google Images; these images have good quality with white background. The total number of images collected by this method is 4,700 with 400-700 images for each category. After removing low quality images, the number of images used for experiments is 3,951. Finally, Dataset 1 has 4,031 images. Each garbage category has from 450 to 530 images.
	
	\item {Dataset 2}: This dataset is created manually. To increase the diversity in the classification dataset, we have also taken daily trash pictures, especially some common categories in schools such as tissues, styrofoam boxes, food packets, and so on. The data collected manually has 987 images. Each garbage category has from 95 to 140 images.
\end{itemize}
 \vspace{-0.8cm}
\section{Trash Bin System Architecture} 
\begin{figure}[h]
	\centerline{\includegraphics [width=8.75cm] {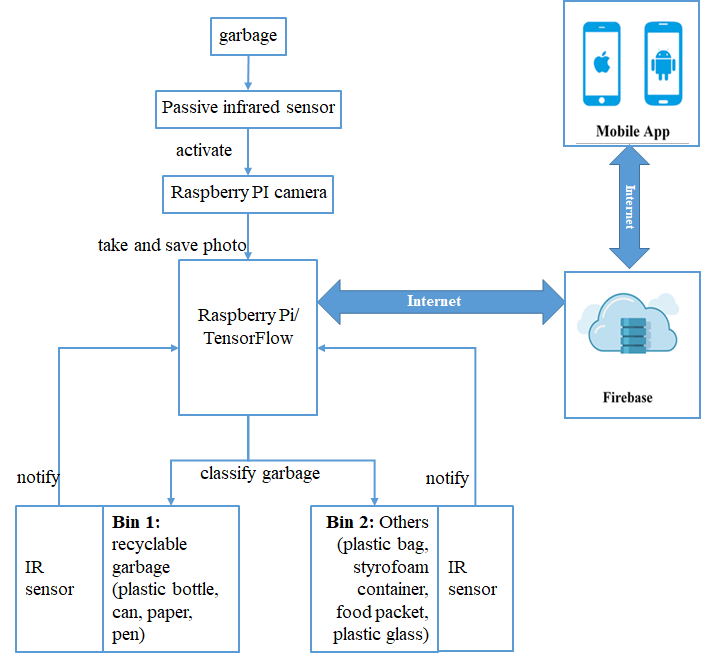}}
	\caption{The trash bin system architecture} \label{fig1}
\end{figure}

\vspace{-0.3cm}
As mentioned earlier, we divide garbage into 2 types: recyclable and non-recyclable trash. The first trash bin, Bin 1, is for the recyclable garbage; the second trash bin, Bin 2, is for the non-recyclable garbage. The whole trash bin system architecture is presented in Fig.~\ref{fig1}.
 \vspace{-0.8cm}
 \vspace{-0.5cm}
 
\begin{figure}[h]
	\centerline{\includegraphics [width=8.5cm] 
		 {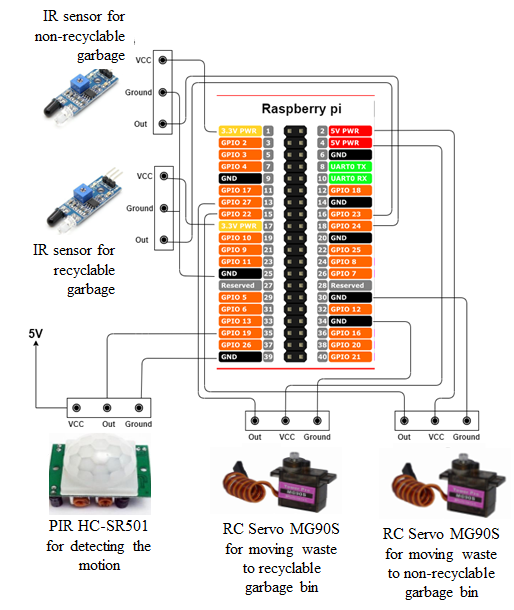}}
	\caption{The circuit diagram of the system} \label{fig2}
\end{figure}

The trash bin system comprises 3 main components connected via the Internet: trash bins running a classification model, a database, and a mobile application. Whenever the trash bin is full, a signal will be sent to the Firestore real-time database and is transferred to the mobile application in real-time mode. The classification processing flow contains the following main steps: When the PIR sensor detects trash, it will wake up the system. Then, the camera is activated for capturing trash and stored temporarily in Raspberry Pi storage. Next, the Raspberry Pi (TensorFlow Lite) uses the captured image for analysis. Immediately after getting the classification result, if the trash is classified into recyclable trash, it will be put into Bin 1; otherwise, it will be put into Bin 2. In addition, each bin has an IR sensor to measure the trash volume in the bin. If the bin is full, the system will send a signal to the mobile application to notify users. Fig.~\ref{fig2} presents the circuit diagram of the system.
 \vspace{-0.5cm}
 \vspace{-0.95cm}
\section{Classifier Training and Data Augmentation} \vspace{-0.2cm}
Considering the trash bin system’s usage and cost, we conclude that the system should better be deployed with small and low-power computers. Therefore, CNN Tensorflow model is firstly picked because of the widely supported tools in transforming the result into Tensorflow Lite model format. MobilenetV2-based Tensorflow Lite results in light-weight and competitive performance to be compatible with a small computer such as Raspberry Pi device, the computer picked based on the system requirement. Moreover, it can also be optimized further for ARM processors on the Raspberry Pi computer. The training process is based on transfer learning, which uses pre-trained models to create a good model for a new task with a small dataset and adding a fully-connected classifier on top.

\subsection{MobileNet V2 Model}
\begin{figure}[h]
	\centerline{\includegraphics [width=\columnwidth]{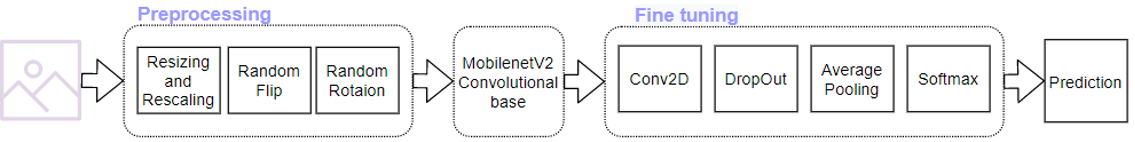}}
	\caption{MobilenetV2-based Tensorflow model} \label{model}
\end{figure}

The base model used in this research is from the MobileNet V2 model \cite{ref_Sandler18}, developed by Google developers with 1.4M images and 1,000 categories. In our implementation, the process to create the base model is provided by TensorFlow\footnote{https://github.com/tensorflow/examples/blob/master/community/en/flowers
	\_tf\_lite.ipynb}, as presented in Fig.~\ref{model}. 
 \vspace{-0.5cm}
 \vspace{-0.5cm}
 
We divide each dataset into two parts consisting of 80\% images for training and 20\% images for validation. The accuracies the model achieves are 85.59\% and 71.00\% on Dataset 1 and Dataset 2, respectively. To improve the accuracy of the classifier, we need to increase the size of the data. In addition, Costa et al. \cite{ref_Costa18} suggest that ``the accuracy in CNN approaches can be improved through some techniques as augmentation, and fine-tuning''. The rest of this section will present methods to augment data to enrich the dataset. \vspace{-0.5cm}

\subsection {Augment Data with OpenCV}\vspace{-0.1cm}
The OpenCV-Python\footnote{https://opencv-python-tutroals.readthedocs.io/en/latest/} provides many tools to transform images. In this study, the Geometric Transformations of Images method is used to extract key features of classes. This method learns to apply different geometric transformations to images, such as translation, rotation, and affine transformation.

The process to augment images is only applied on Dataset 2. The transform process creates 4 new images from one given image. After transforming and removing poor images, this dataset consists of 4,264 images. The accuracy the model achieves on this dataset is 57\%. Somehow this method does not work well on our manually collected dataset. For future work, we will address this issue.\vspace{-0.5cm}

\subsection {Augment Data with Keras Preprocessing Layer }\vspace{-0.2cm}
Another widely used method to augment data is to use the Keras Preprocessing layer\footnote{https://keras.io/guides/}. TensorFlow provides a tf.keras.layers.experimental.preprocessing module. This module includes several layers for preprocessing and augmenting images. We use the \textit{Resizing} and \textit{Rescaling} classes to standardize the input images for the model. We then use \textit{RandomFlip} and \textit{RandomRotation} to augment images and transform images to new images for the training model. The process of augmenting images is performed on 1,984 images including 997 images from Dataset 1 and 987 images from Dataset 2. The accuracy of the classifier on this dataset achieves 90\%. 

With almost 1.000 images collected manually from real life garbage, the experimental trash bin product shows a considerably good result for a small real-life-sample. The test result achieves 86.67\% accuracy on 30 popular trash pieces. The rest 13.33\% failures are happened on crumpled plastic bags and papers because of their similar visual texture. Besides, some unexpected samples can lead to unknown results, for example, a phone charger is categorized into recyclable garbage. However, these problems can be minimized with further improvement on the training dataset. \vspace{-0.5cm}

\section{Implement Classifier on Raspberry Pi}\vspace{-0.3cm}
Since a Raspberry Pi could be considered as a mini-computer, running a program on Raspberry Pi is quite similar to running on a regular computer. First, we installed the necessary libraries for the program. Although the interpreter could be directly used from the TensorFlow Lite library, the number of documentations for installing the Lite version on Raspberry was still limited, which caused difficulty during the implementation process. Therefore, we installed the entire the TensorFlow library, which contains the TensorFlow Lite library. Then, the trash classifier will be loaded into the Raspberry Pi. After the classifier has been successfully run on Raspberry, connecting the Raspberry Pi with other support devices will help the trash bin run automatically.
\begin{itemize}
	\item Raspberry Pi camera module: The \textit{picamera} and \textit{time} libraries are used to manage the camera.
	\item PIR sensor: This sensor is used to detect people coming near the trash bin. The \textit{RPi.GPIO} library is imported to manage the pins.
	\item Servo motors: Two motors are responsible for rotating a temporary garbage container to the correct bin and dropping the garbage. The \textit{ChangeDutyCycle} library is used to control the rotation of the motors.  
\end{itemize}
\vspace{-0.7cm}
\section{Mobile Application for Managing Trash Bin}\vspace{-0.3cm}
A smart trash management application will help manage the trash system effectively. Our application is built based on React-Native; so, it is compatible with smartphones running Android and iOS. Besides, the application can update the trash bin status in real-time by using the Firebase real-time database\footnote{https://firebase.google.com/} and send messages to notify users whenever the trash bin is full. Each trash bin is identified by a unique identification number (\textit{ID}), creation date (\textit{date}), location of the bin (\textit{locate}), bin status (\textit{status}), bin description (\textit{description}), and an image of the bin (\textit{image}). The mobile application provides users with 3 main services:
\begin{itemize}
	\item Display: The application gets information from the database and shows on the screen information for users to observe, including the status (full or normal) of the trash bin.
	
	\item Management: The application allows users to add or remove trash bins to their working list to view or update the information about bins.
	
	\item Notification: Whenever bins are full, users will get notifications.	
\end{itemize}

\vspace{-0.7cm}
\section {Conclusion}\vspace{-0.3cm}
We propose a hardware implementation of a system that automatically classifies trash bin contents with good results. By using a machine learning model and IoT's automobility, the trash bin can help simplify the process to classify domestic waste: users need only to separate waste into single items and put them on the trash bin plate, it will classify itself. The mobile application can also support users in monitoring the filled status of the bin and notify them to clean it. This use case is especially useful when the system is deployed in public spaces, where it is not effective to check each trash bin status manually.
We plan to deploy the smart trash bin system at the College of Information and Communication Technology at Can Tho University in Vietnam. To create a more accurate system, we will increase the size of the trash image dataset and study methods to augment the collected data. Currently, the system can classify only one type of waste at a time. In the future, we will study methods to classify many types of garbage. 
\end{document}